\documentclass[letterpaper]{article} 
\usepackage{aaai25}  
\usepackage{times}  
\usepackage{helvet}  
\usepackage{courier}  
\usepackage[hyphens]{url}  
\usepackage{graphicx} 
\usepackage{natbib}  
\usepackage{caption} 
\usepackage{algorithm}
\usepackage{algorithmic}
\usepackage{amsmath,amssymb}
\usepackage{amsmath}
\usepackage{bm}
\usepackage{booktabs}
\usepackage{newfloat}
\usepackage{listings}
\DeclareCaptionStyle{ruled}{labelfont=normalfont,labelsep=colon,strut=off} 
\floatstyle{ruled}
\newfloat{listing}{tb}{lst}{}
\floatname{listing}{Listing}
\title{Fast and Interpretable Mixed-Integer Linear Program Solving \\ by Learning Model Reduction}
\author{
    Yixuan Li\textsuperscript{\rm 1}, Can Chen\textsuperscript{\rm 1}, Jiajun Li\textsuperscript{\rm 1}, Jiahui Duan\textsuperscript{\rm 2}, Xiongwei Han\textsuperscript{\rm 2}, Tao Zhong\textsuperscript{\rm 2}, Vincent Chau\textsuperscript{\rm 1}, Weiwei Wu\textsuperscript{\rm 1}, Wanyuan Wang\textsuperscript{\rm 1}\thanks{Corresponding author}}
\affiliations{
    \textsuperscript{\rm 1}School of Computer Science and
Engineering, Southeast University\\
    \textsuperscript{\rm 2}Noah's Ark Lab, Huawei Technologies\\\{yixuanli,chencan0,jiajun\_li,vincentchau,weiweiwu,wywang\}@seu.edu.cn, \{jiahui.duan,xiongwei,zhongtao5\}@huawei.com
}

\usepackage{bibentry}
\usepackage{multirow}

\begin{document}

\maketitle

\begin{abstract}
By exploiting the correlation between the structure and the solution of Mixed-Integer Linear Programming (MILP), Machine Learning (ML) has become a promising method for solving large-scale MILP problems. Existing ML-based MILP solvers mainly focus on end-to-end solution learning, which suffers from the scalability issue due to the high dimensionality of the solution space. Instead of directly learning the optimal solution, this paper aims to learn a reduced and equivalent model of the original MILP as an intermediate step. The reduced model often corresponds to interpretable operations and is much simpler, enabling us to solve large-scale MILP problems much faster than existing commercial solvers. However, current approaches rely only on the optimal reduced model, overlooking the significant preference information of all reduced models. To address this issue, this paper proposes a preference-based model reduction learning method, which considers the relative performance (i.e., objective cost and constraint feasibility) of all reduced models on each MILP instance as preferences. We also introduce an attention mechanism to capture and represent preference information, which helps improve the performance of model reduction learning tasks. Moreover, we propose a {\sc SetCover} based pruning method to control the number of reduced models (i.e., labels), thereby simplifying the learning process. Evaluation on real-world MILP problems shows that 1) compared to the state-of-the-art model reduction ML methods, our method obtains nearly 20\% improvement on solution accuracy, and 2) compared to the commercial solver Gurobi, two to four orders of magnitude speedups are achieved. 
\end{abstract}
%

\setlength{\floatsep}{8pt}  
\setlength{\textfloatsep}{8pt}  
\setlength{\intextsep}{8pt}  
\setlength{\dblfloatsep}{8pt}     
\setlength{\dbltextfloatsep}{8pt} 

\section{Introduction}
Due to its strong expressiveness, Mixed-Integer Linear Programming (MILP) has been widely used in various critical domains, including supply chain and logistics \cite{Chao2024}, service scheduling \cite{RosemarinRK19}, energy management \cite{6485014}, transportation planning \cite{LowalekarVJ21,li2024factor}, chip design \cite{wangcircuit,wang2024hierarchical}, and chemistry research \cite{gengnovo}. Commercial solvers, such as Gurobi, Cplex and Matlab, are mainly used to solve MILP problems. In real-world industrial applications, MILP instances often involve hundreds of thousands of decision variables and constraints \cite{6485014,Li2021}. Existing commercial solvers are based on exact solutions, which are computationally expensive and cannot meet the real-time demands of industrial applications. 


In many scenarios, a large number of homogeneous MILPs with similar combinatorial structures need to be solved simultaneously. For example, online stochastic programming often involves solving similar MILP instances at each stage, with slightly modified input parameters while the structure remains unchanged \cite{LowalekarVJ18,bertsimas2022online}. Machine Learning (ML), with its powerful pattern recognition capability, can exploit the correlation between the structure and the solution of MILP, and has recently become a very promising research topic for solving large-scale MILP \cite{BengioLP21,ZhangLLZYLY23,HentenryckD24}. Existing ML-based MILP solvers can be classified into two categories: 1) end-to-end solution prediction, i.e., directly learning the mapping between MILP instances and solutions \cite{DontiRK21,DingZSLWXS20,Chen0WY23a}; and 2) learning to optimize, i.e., learning to improve the process of traditional solvers \cite{HeDE14,KhalilDZDS17,SongLYD20,ChiAKWS22,LingWW24,HanYCZZW0L23,BalcanDSV24}. Due to the high dimensionality of the solution space, existing ML methods that learn the optimal solution as a function of the input parameters, suffer from the scalability issue. Furthermore, it is currently not possible to interpret the predicted solution or to understand it intuitively  \cite{ParkH23}.  


Instead of directly learning the optimal solution, this paper takes a different method to learn a reduced and equivalent model of the original MILP as an intermediate step. In Operations Research \cite{BV2014}, an equivalently reduced model of the MILP constitutes the minimal information, including the set of active constraints and the value of integer variables at the optimal solution, required to recover the optimal solution. Model reduction learning has the following three advantages \cite{MisraRN22}: 1) from the optimization perspective, the reduced model is much easier than the original MILP model, which can be solved fast, 2) from the ML perspective, using the reduced models as labels can reduce the dimension of the learning task, and 3) from the application perspective, the reduced model often corresponds to interpretable modes of operation, which can assist human engineers in decision making \cite{bertsimas2021voice}. 

To the best of our knowledge, \cite{MisraRN22,BertsimasK23} are the only works that learn model reduction for fast MILP solving. Their idea is to train a classification algorithm that aims to predict the correct label (i.e., reduced model). However, the algorithm treats a set of feasible reduced models as equally desirable labels, failing to fully exploit the comparative information available in the reduced model space. To tackle this challenge, this paper transforms the performance (i.e., objective function value and constraint feasibility) of a reduced model on an MILP instance as preferences, and proposes a preference-based reduced model learning method. An attention-based encoder then utilizes the ranked preferences is proposed to extract correlations between instances and reduced models. 

The contributions of this paper are summarized as follows. First, we introduce a different model reduction approach to learn the optimal solution of MILP problems. To improve learning accuracy, we fully exploit the preference information in terms of the performance of reduced models on instances. We also integrate an attention architecture and preference-based loss function to capture the correlations between instances and the preferred reduced models. Second, to avoid the number of labels (i.e., reduced models) growing quickly with the number of instances, we propose a { \sc SetCover} technique to generate the minimum labels that are feasible for all instances. Finally, we conduct extensive experiments on real-world domains to validate the proposed preference-based model reduction learning method. Results show that 1) our method can prune redundant reduced models efficiently, and 2) our method has a significant improvement in finding accurate solutions within seconds.

\section{Related Work}

Existing ML-based MILP solving methods can be categorized into three groups: 1) end-to-end solution prediction, i.e., directly learning the mapping between MILP instances and solutions; 2) learning to optimize, i.e., learning to accelerate the solving process of traditional exact/heuristic methods; 3) learning to simplify the MILP, i.e., learning to pre-solve or reduce the size of the MILP formulation.

\textbf{End-to-end Solution Prediction.} Using ML to learn the mapping from MILP instances to a high-dimensional solution space is straightforward, however, it often results in low prediction accuracy \cite{DontiRK21,ParkH23,Chen0WY23a}. Therefore, \cite{nair2020solving,ye2023gnn} only predicts values for partial variables and computes the values of the remaining variables using the off-the-shelf solver. Directly predicting variable values cannot maintain the hard constraints \cite{DingZSLWXS20}. Instead, \cite{HanYCZZW0L23} predicts an initial solution and searches for feasible solutions in its neighborhood. 

\textbf{Learning to Optimize.} For exact solving, there are always hyperparameters and selection rules that need to be fine-tuned to accelerate the solving process. For example, the selection of branching variables and their values in Branch-and-Bound, the selection of the cutting rules in Cutting Plane, and the column generated in the Column Generation algorithm. Using experienced data of these exact solvers, Imitation Learning (IL) and Reinforcement Learning (RL) have been used to learn effective hyperparameters and selection rules \cite{wang2024asp,wangHEM,huang2022branch,lin2022learning,wang2023learning}. On the other hand, for heuristic algorithms such as Local Search Heuristics \cite{cai2013local}, and Large Neighborhood Search Heuristics \cite{song2020general,wu2021learning}, ML can also be used to improve their effectiveness. For example, \cite{qi2021smart} use RL to iteratively explore better solutions in Feasible Pump, and \cite{nair2020neural} use RL to search for better solutions within a neighborhood.

Although these two directions introduce MILP to the benefits of ML and show promising results, they do not scale well to real-world applications. Directly predicting a high-dimensional solution is intractable. The efficiency of IL and RL-based optimization methods is limited by the decision horizon \cite{ye2024state,geng2024reinforcement,wang2022sample} (i.e., the number of integer variables). Another drawback is their inability to enforce the constraints accurately, making them unsuitable for real-world high stakes applications \cite{liu2023promoting,liumilp,geng2023deep}. In contrast, this paper utilizes model reduction theory and focuses on learning the mapping between an MILP instance and its optimal reduced model, providing a fast and interpretable MILP solution. 


\textbf{Learning to Simplify the MILP.} Large-scale MILP formulations usually contain much redundancy, which can be simplified by the pre-solve techniques \cite{achterberg2013mixed}. To design high-quality pre-solve routines, \cite{liul2p} and \cite{kuang2023accelerate} recently use RL to determine which pre-solve operators to select and in what order. \cite{ye2023light} instead use graph partition for problem division to reduce computational cost. These pre-solve-based simplification methods can only identify limited and explicit redundancy. To identify the minimal tight model, \cite{MisraRN22,BertsimasK23} first propose a classification method to predict the optimal reduced model. However, existing methods only consider several equally desirable reduced models, overlooking the various performances of the reduced models. This paper considers the importance of preference information and designs an efficient method to exploit it to improve the learning accuracy of model reduction.    
\begin{figure*}[t]
\centering
\includegraphics[width=0.9\textwidth]{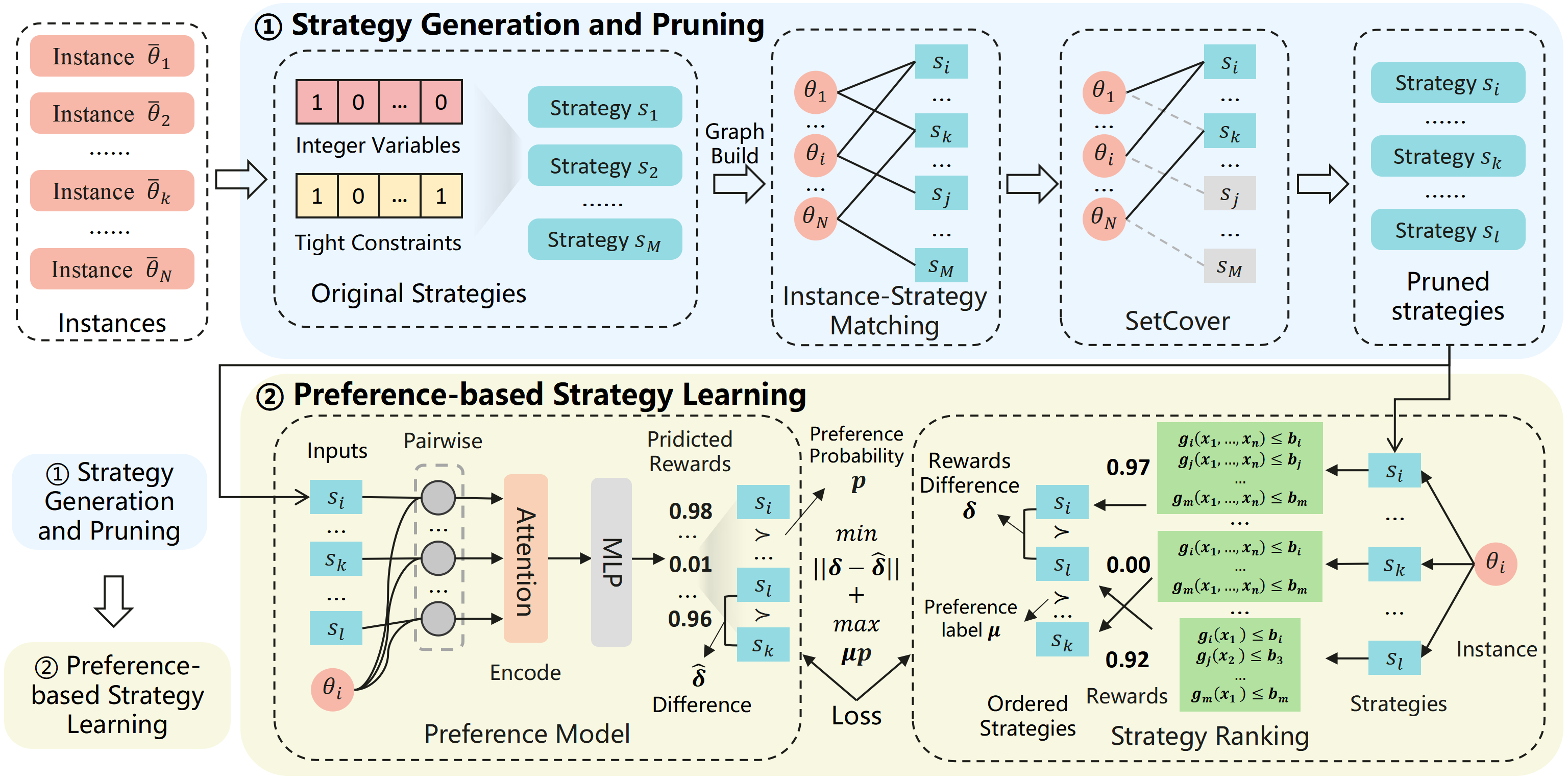} 
\caption{Overview of our framework, which comprises two phases: 1) \textit{Strategy Generation and Pruning} and 2) \textit{Preference-based Strategy Learning}. In 1), the strategies are explored from instances and a {\sc SetCover} is constructed to prune redundant strategies. In 2), an attention architecture is utilized to capture the ranked preference information over instances.} 
\label{fig_overall}
\end{figure*}
\section{Parameterized MILP and Model Reduction}
\textbf{Parameterized MILP Problem.} The MILP can be formalized as follows:
\begin{align}
    &\min_{x} \; f(c, x)
\end{align}
\begin{equation}
\begin{aligned}
&\text{s.t.} \; ~~~g(A, x) =
\begin{cases}
g_1(A_1, x_I,x_{-I}) \leq b_1 \\
g_2(A_2, x_I,x_{-I}) \leq b_2 \\
\vdots \\
g_m(A_m, x_I,x_{-I}) \leq b_m
\end{cases}
\end{aligned}
\end{equation}
\begin{align}
\quad x_I \in \mathbb{Z}^d, \quad x_{-I} \in \mathbb{R}^{n-d}
\end{align}

\noindent where $c \in \mathbb{R}^{n}$ denotes the coefficient vector of the objective function, $f: \mathbb{R}^{n} \times \mathbb{R}^{n} \to \mathbb{R}$ is the objective function. $g_i: \mathbb{R}^{n} \times \mathbb{R}^{n} \to \mathbb{R}$ denotes the $i$-th constraint and $A_i$ denotes its coefficients. $b \in \mathbb{R}^{m}$ denotes the parameter of these $m$ constraints.  The $x \in \mathbb{R}^{n}$ is $n$ decision variables, and $I (|I| = d)$ is the set of integer decision variables, i.e., $x_{I} \in \mathbb{Z}^{d}$. Let $\theta = \langle A,c,b \rangle$ denote the parameters of the MILP problem, each parameter represents a specific MILP instance. 

\textbf{The Strategy of Model Reduction.} Given a parameter $\theta$, let $x^{*}(\theta)$ denote the optimal solution.  We denote the \textit{tight constraints} $\mathcal{T}(\theta)$ are constraints that are equalities at the optimal solution $x^{*}(\theta)$, i.e., 
\begin{equation} 
    \mathcal{T}(\theta) = \{i \in \{ 1,...,m\}|g_{i}(A_i,x^{*}(\theta))=b_i\}.
\end{equation}
Given $\mathcal{T}(\theta)$, all other constraints in the MILP model are redundant and can be removed. For MILP problems, the number of tight constraints is at most $n$ (i.e., $|\mathcal{T}(\theta)| \leq n$) \cite{NW2006}. Since some components of $x$ are integers, it is still not trivial to compute the optimal solution by only knowing $\mathcal{T}(\theta)$. By fixing the integer components to their optimal values $x^{*}(\theta)$, the tight constraints allow us to efficiently compute the optimal solution. We now define the optimal \textit{strategy} of model reduction as the set of tight constraints together with the value of integer variables at the optimal solution, i.e., $s^{*}(\theta)=(\mathcal{T}(\theta),x^{*}(\theta))$. Finally, given the optimal strategy $s^{*}(\theta)$, the original MILP model (1)-(3) can be reduced to a LP model:
\begin{align}
    &\min_{x} \; f(c, x)
\end{align}
\begin{equation}
\begin{aligned}
&\text{s.t.} \; g_i(A_i, x_I,x_{-I}) \leq b_i, \forall i \in \mathcal{T}(\theta) \\
\end{aligned}
\end{equation}
\begin{align}
\quad x_I = x^{*}_{I}(\theta), \quad x_{-I} \in \mathbb{R}^{n-d}
\end{align} 
The reduced LP model (5)-(7) is much easier because LP is continuous and has a smaller number of constraints.

\textbf{Objective.} Given an MILP problem, although different MILP instances are not the same, only the key parameters vary slightly, and the structure remains unchanged. We can exploit the repetitive structure of the MILP instances and solutions and learn to solve unseen MILP instances. Therefore, our objective in this paper is to learn the mapping from the parameter $\theta$ to the optimal strategies $s^{*}(\theta)$ as an intermediate step for fast MILP solving. 
\section{Framework Overview}
The proposed framework comprises two phases: 1) \textit{Strategy Generation and Pruning} for exploring a set of useful strategies $S$ as labels, and 2) \textit{Preference-based Strategy Learning} for predicting the correct strategy $s \in S$ for any instance $\theta$ (see Figure~\ref{fig_overall} for details). 

\textit{Strategy Generation and Pruning} aims to generate a set of useful strategies that can be applied to all MILP instances. Intuitively, we can explore as many MILP instances $\theta$ as possible and denote their optimal strategy $s^{*}(\theta)$ as  candidate strategy labels. However, the number of candidate strategies can grow quickly with respect to the number of instances, thereby making the strategy learning task very difficult. This paper proposes a {\sc SetCover} technique for strategy pruning while ensuring that all generated instances can be covered, in which the \textit{cover} means that there is at least one feasible strategy for an instance. 



\textit{Preference-based Strategy Learning} aims to propose a machine learning approach to predict the optimal strategy $s^{*}(\theta)$ for any MILP instance $\theta$. Given the training data $\left \{ \theta_{i}, s^{*}(\theta_{i})\right \} $, existing strategy learning approaches typically only focus on the exact optimal strategy  \cite{MisraRN22} or treat a set of feasible strategies equally desirable \cite{BertsimasK23}, overlooking the significant preference information available in the instance-strategy space. To address this issue, this paper transforms the performance (i.e., objective function value and constraint feasibility) of strategies on MILP instances as preferences, and proposes a novel preference-based strategy learning method. 


\section{Strategy Generation and Pruning}
In this section, our objective is to identify a set of useful strategies that will be used as labels for model reduction training and learning. 

\textbf{Strategy Generation.} It is difficult to determine the amount of instances required for strategy learning. Following the approach in \cite{bertsimas2022online}, we first randomly generate instances $\theta$ as well as their optimal strategy $s^{*}(\theta)$ until the Good-Turning estimator $\frac{N_1}{N}$ falls below a tiny value, where $N$ is the total number of instances and $N_{1}$ is the number of different strategies appeared exactly once. Specially, given $N$ independent instances $\Theta_{N}=\{\theta_1,\cdots,\theta_N\}$, we can generate $M$ different strategies $S(\Theta_{N})=\{s_1,\cdots,s_{M}\}$. However, in large-scale MILP problems, the number of strategies (i.e., labels) $M$ can grow quickly, making the learning task very difficult. Motivated by this issue, we next propose a strategy pruning method based on {\sc SetCover} technique. 

\textbf{Strategy Pruning.} In practice, each optimal strategy $s^{*}(\theta_{i})$ not only applies to the corresponding instance $\theta_{i}$, but also may apply to other instances $\theta_{j} (\neq \theta_{i})$. Therefore, many candidate strategies are redundant and we can select only the most useful strategies to apply. We first model the relationship between strategies $S(\Theta_{N})$ and instances $\Theta_{N}$ by an Instance-Strategy bipartite graph $G(V_\theta,V_s,E)$:
\begin{itemize}
\item The node set consists of the instance nodes $V_\theta = \{v_\theta^1,...,v_\theta^N\}$ and the strategy nodes $V_s = \{v_s^1,...,v_s^M\}$. Each  $v_\theta^i$ represents an instance $\theta_i\in \Theta_{N}$, and each  $v_s^j$ represents a strategy $s_j\in S(\Theta_{N})$. 
\item An edge $e_{i,j}\in E$ exists between  $v_\theta^i$ and  $v_s^j$ if applying the strategy $s_j$ to the instance $\theta_i$, and the infeasibility $p(\theta_i,s_j)$ and suboptimality $d(\theta_i,s_j)$ of the reduced problem (5)-(7) are both below a tiny threshold $[\epsilon_p,\epsilon_d]$. The infeasibility is defined as:
\begin{align}
\label{eqinfeas}
p(\theta_i,s_j) = \|(g(A, \hat{x}_{i,j}^*)-b)_+ \|_\infty/\|b\|,
\end{align}
where $\hat{x}_{i,j}^*$ is the solution of the reduced  problem\footnote{The way we apply a strategy to a MILP instance is to fix the integer variables to the values specified in the strategy, impose only tight constraints, and solve the resulting LP problem.}.  $||b||$ normalizes the degree of constraint violation based on the magnitude of the constraint $g(A,x) \le b$. The Suboptimality measures the relative distance between the recovered solution $\hat{x}_{i,j}^*$ of the reduced problem and the optimal solution $x_i^*$ of the instance $\theta_i$:
\begin{align}
\label{eqsub}
d(\theta_i,s_j) = |f(c,\hat{x}_{i,j}^*) - f(c,x_{i,j}^*)|/ |f(c,x_{i,j}^*)|.
\end{align}
\end{itemize}

The objective of strategy pruning is to find a minimal subset of strategy nodes $V_s^{'} \subseteq V_s$, such that for any $v_\theta^i \in V_\theta$, there exists a $v_s^j \in V_s^{'}$ and $e_{i,j} \in E$. This problem of strategy pruning can be reduced to the well known {\sc SetCover} problem of finding the minimum  sets to cover all elements. Thus, an efficient greedy algorithm can be employed to find the useful strategies \cite{KHULLER199939}. The main idea of the greedy algorithm is to iteratively select the strategy node $v_s^*$ that is connected to the maximal uncovered instance nodes. This node $v_s^*$ is then added to the candidate set of strategies $V_s^{'}$ and this strategy selection process continues until all instance nodes are connected to at least one candidate strategy node. Finally, the set of pruned strategies $S^P$ can be obtained from $V_s^{'}$.

\section{Preference-based Strategy Learning}
Given the pruned strategies $S^P$, this section proposes to learn the mapping from a MILP instance $\theta$ to a suitable strategy $s(\theta)\in S^P$. Previous strategy learning approaches \cite{MisraRN22,bertsimas2022online} focus on predicting the correct strategy and considering all other strategies equally undesirable, failing to integrate the significant instance-strategy preference information deeply. 

\textbf{Preference Computation.} Given an instance $\theta_i$, we would like to supply a reward $r(\theta_i, s_j)$ to each strategy $s_j \in S^P$. The reward $r(\theta_i, s_j) \in r$ is used to measure the outcome of applying the strategy $s_j$ to the instance $\theta_i$. We follow the same criteria for calculating the relative feasibility and suboptimality as in Eqs. (\ref{eqinfeas}) and (\ref{eqsub}):
\begin{align} \label{Eq:RF}
r(\theta_i, s_j) = -log(p(\theta_i, s_j) + d(\theta_i, s_j)).  
\end{align}
Directly learning the real reward function $r(\theta_i, s_j)$ between $\theta_i$ and $s_j$ is extremely challenging because the complex relationships among instances, strategies and rewards. Instead, to enhance simplicity and training stability, we propose to learn a proxy reward model $R_{\phi}$ (where $\phi$ denotes the parameters of the machine learning approach) that can express {\it preferences} between strategies. Given an instance $\theta_{i}$, for two instance-strategy pairs \((\theta_i, s_j)\) and \((\theta_i, s_k)\), we define the preference $\succ$ generated by the rewards $r$:
\begin{align} 
\label{eq:preference}
(\theta_i, s_j) \succ  (\theta_i, s_k) \; \Leftrightarrow \; r(\theta_i, s_j) > r(\theta_i, s_k).
\end{align}
Informally, $(\theta_i, s_j) \succ  (\theta_i, s_k)$ indicates that the instance $\theta_{i}$ prefers the strategy $s_j$ to the strategy $s_k$. 







\begin{algorithm}[tb]
\caption{Preference-based Strategy Learning}
\label{alg:2}
\textbf{Input}: Training data \(\{[\theta_i, S^P] = [\langle\theta_i, s_1\rangle,...,\langle\theta_i, s_{M^P}\rangle] \mid \theta_i \in \Theta_{N}\}\), rewards $r = \{r(\theta_i,s_j) \mid \theta_i \in \Theta_{N},s_j \in S^P\}$, preference model $R_\phi$, learning rate \(\alpha\), weight $\lambda_1$ and $\lambda_2$. \\
\textbf{Output}: Optimized model parameters $R_\phi$.

\begin{algorithmic}[1] 
\STATE \textbf{Initialize} model parameters $R_\phi$.
    \FOR{each instance \(\theta_i\)}
        \STATE Predict rewards \(\{\hat{r}_{i,1},...,\hat{r}_{i,M^P}\} = R_\phi([\theta_i, S^P])\).
        \STATE Rank strategies $S_\sigma^P=\{s_{\sigma(1)},...,s_{\sigma(M^P)}\}$ by rewards $r$ to get predicted rewards $\{\hat{r}_{i,\sigma(1)},...,\hat{r}_{i,\sigma(M^P)}\}$.
        \FOR{adjacent strategies $s_{\sigma(j)},s_{\sigma(j+1)}$ in $S_\sigma^P$}
            \STATE \(p_{i,j} = \frac{exp(\hat{r}_{i,\sigma(j)})}{exp(\hat{r}_{i,\sigma(j)})+exp(\hat{r}_{i,\sigma(j+1)})}\).
            \STATE Compute difference $\hat{\delta}_{i,j}=\hat{r}_{i,\sigma(j)}-\hat{r}_{i,\sigma(j+1)}$.
        \ENDFOR
        \STATE Compute Preference Loss $L_p(\phi)$ by $p_{i,*}$ and $S_\sigma^P$.
        \STATE Compute MSE Loss $L_d(\phi)$ by $\hat{\delta}_{i,*}$ and rewards $r$.
    \ENDFOR
        \STATE Compute Total Loss $L_{total}(\phi) = \lambda_1L_p(\phi) + \lambda_2L_d(\phi)$ 
        \STATE Update model parameters \(\phi \leftarrow \phi - \alpha \nabla_\phi L_{total}(\phi)\).
\STATE \textbf{Return} optimized parameterized model $R_\phi$.
\end{algorithmic}
\end{algorithm}

\textbf{Preference-based Sampling.} To train the proxy reward model $R_{\phi}$, previous preference-based learning (e.g., RLHF \cite{Christiano17}) requires selecting all possible pairwise comparisons as samples. For example, let $M^{P}=|S^P|$ denote the number of pruned strategies, there will be  $\binom{M^{P}}{2}$ preference samples for each instance at the training stage. The number of samples grow quadratically with the number of strategies, thereby increasing the cost of training. Fortunately, in our strategy learning problem, the instance preferences on strategies have a transitivity structure. For example, given the instance $\theta_i$, if the strategy $s_j$ is preferred to $s_k$, and $s_k$ is preferred to $s_q$, we still have that $s_j$ is preferred to $s_q$. This transitivity property can rank all candidate strategies as a complete order based on preferences (i.e., Eq. (\ref{eq:preference})). Therefore, for each instance $\theta_{i}$, when the candidate strategies are ranked in decreasing order of their preferences (i.e., $s_{\sigma(j)} \in S_\sigma^P=\{s_{\sigma(1)},...,s_{\sigma(M^P)}\}$ is ranked by its reward $r(\theta_i,s_{\sigma(j)})$, and $\sigma(j)$ represents the new position of $s_{\sigma(j)}$ in the sequence $r(\theta_i,s_{\sigma(1)})\geq r(\theta_i,s_{\sigma(2)})\geq \cdots r(\theta_i,s_{\sigma(M^P)}) $), only the $M^{P}$ ordered preference samples are necessary as the preference set $\mathcal{D}(\theta_i)$ for $\theta_i$:
\vspace{-2pt}
\begin{align}  
\{\langle(\theta_i, s_{\sigma(1)}) \hspace{-3pt}\succ\hspace{-3pt}  (\theta_i, s_{\sigma(2)})\rangle,\cdots,\succ\hspace{-3pt}  (\theta_i, s_{\sigma (M^P)})\rangle\}.
\end{align}
The size of the samples in $\mathcal{D}(\theta_i)$ increases linearly with the number of strategies, avoiding a large amount of redundant preference samples in $\binom{M^{P}}{2}$.

\textbf{Attention-based Instance-Strategy Encoding.} An attention architecture is proposed to improve the representation capacity. The input to the architecture is a vector of instance-strategy pairs, $[\theta_i, S^P]=[\langle\theta_i, s_1\rangle,\cdots,\langle\theta_i, s_{M^P}\rangle]$, and the output is a vector of the predicted rewards $\hat{R}_{i}=\{\hat{r}_{i,j}\}_{j=1}^{M^P}$, where $\hat{r}_{i,j}$ is the predicted reward for strategy $s_j$ applied on instance $\theta_i$. To extract the inherent similarity among strategies as well as the underlying connections between instances and strategies, we apply an attention mechanism to encode instance-strategy pairs. Specifically, we treat each instance-strategy pair $\langle\theta_i, s_j\rangle$ as a token, allowing all pairs $[\theta_i, S^P]$ to be considered when encoding $\langle\theta_i, s_j\rangle$. This architecture can prioritize the more important pairs and extract features of strategies that can effectively solve instances. The above process can be expressed by the following formula: 
\vspace{-2pt}
\begin{align}
\label{eq:model}
\textstyle A([\theta_i, S^P]) = softmax\left( \frac{QK^T}{\sqrt{d}}\right)V.
\end{align}
In Eq. (\ref{eq:model}), based on the row-wise shared weights $W^q$, $W^k$, and $W^v$, a linear projection operation is acted on the input $[\theta_i, S^P] $ to compute the queries $Q = [\theta_i, S^P]W^q$, keys $K = [\theta_i, S^P]W^k$, and values $V = [\theta_i, S^P]W^v$. The main architecture consists of $L$ layers:  
\begin{align}
\label{eq:model2}
A_L(A_{L-1}(...A_1([\theta_i, S^P]))).
\end{align}
The final output layer performs an affine transformation:  
\vspace{-2pt}
\begin{align}
\label{eq:model3}
\hat{R}_{i} & = R_\phi([\theta_i, S^P]) = y_L(A_{L}) = \psi_L(W_LA_L+b_L),
\end{align}
\noindent where $\psi_L$ is the activation function, $W_L$ and $b_L$ are weights of the output layer $L$.  $\hat{R}_{i}=\{\hat{r}_{i,j}\}_{j=1}^{M^P}$ represents the predicted rewards of all strategies $s_j \in S^P$ when applied on the instance $\theta_i$. For this architecture, $\phi$ is the whole set of parameters that needs to be learned.

\begin{figure}[t]
\centering
\includegraphics[width=1\columnwidth]{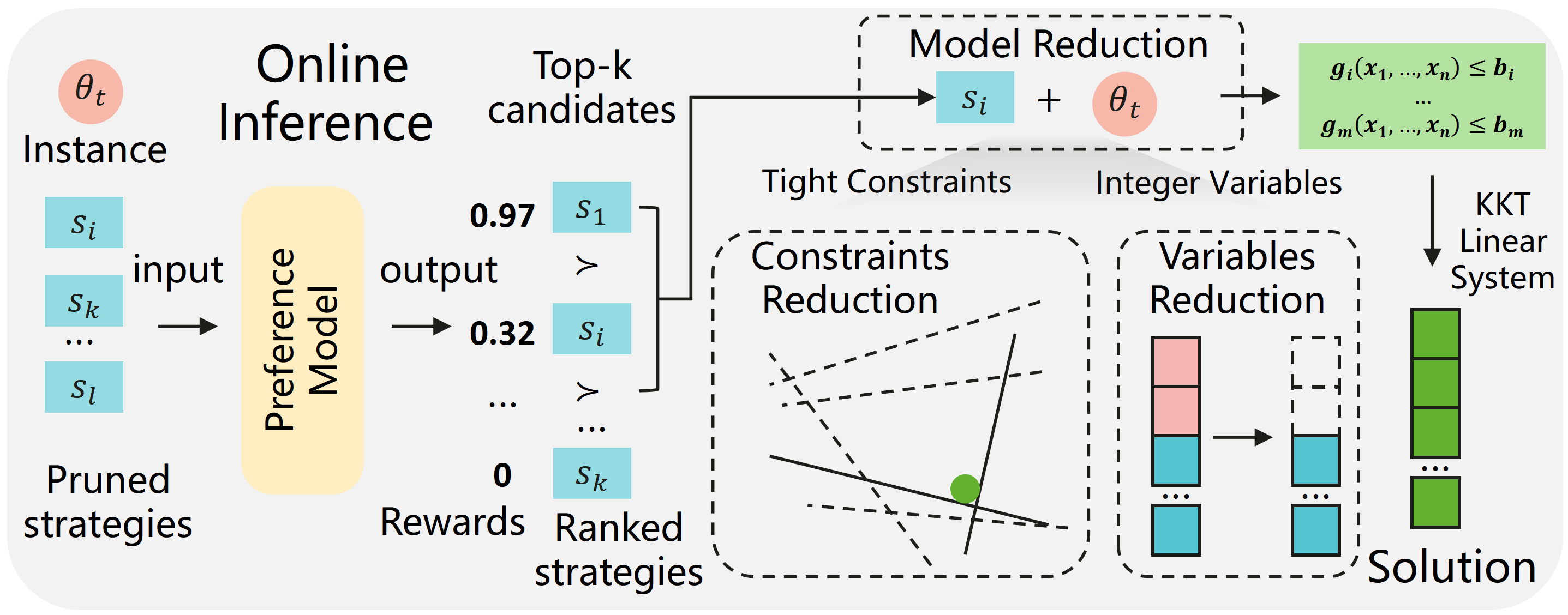}
\caption{The $k$ highest outputs of preference model are selected as candidate strategies. Given the strategy, the instance can be solved rapidly by model reduction.} 
\label{fig_online}
\end{figure}
\begin{figure*}[t]
\centering
\includegraphics[width=0.9\textwidth]{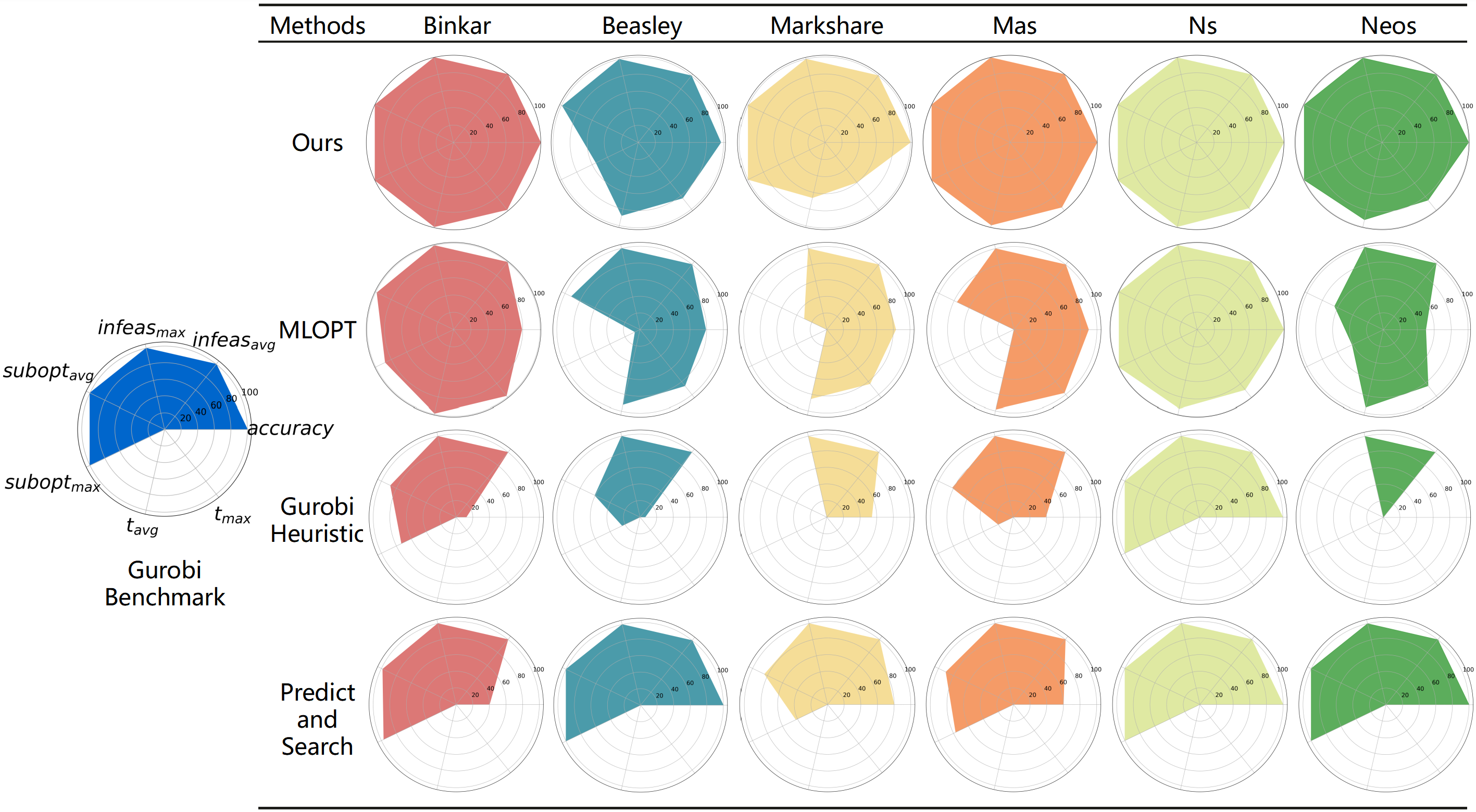} 
\caption{The performance on six scenarios from MIPLIB and each vertex in the subplot represents a metric. For better presentation of results (due to the differences in metric magnitudes), we map each metric to the range of (0, 100) using the same function for each metric. The Gurobi's result is shown on the left since its results are similarly presented across all scenarios} 
\label{radar}
\end{figure*}
\textbf{Preference-based Loss Function.} 
In the training phase, the strategies $S^P$ can be ranked to $S_\sigma^P$ by the rewards $r$ , so we can get the reward output $\hat{R}_{i}$ ordered by $S_\sigma^P$ and form the ranked predicted rewards $\hat{R}_{i,\sigma}=\{\hat{r}_{i,\sigma(1)},...,\hat{r}_{i,\sigma(M^P)}\}$, where $\sigma(j)$ represents the new position of $\hat{r}_{i,\sigma(j)}$ in the sequence. We can define the preference probability $p_{i,j}$ between each pair of adjacent strategies \(s_{\sigma(j)}\) and \(s_{\sigma(j+1)}\):
\vspace{-2pt}
\begin{align}
\label{eq:prob}
p_{i,j} = \frac{\exp(\hat{r}_{i,\sigma(j)})}{\exp(\hat{r}_{i,\sigma(j)}) + \exp(\hat{r}_{i,\sigma(j+1)})}.
\end{align}
\vspace{-2pt}
In order for the model output $\hat{R}_{i}$ to yield correct relative preferences, we need to train $R_\phi$ to maximize the probability $p_{i,j}$ for every ordered adjacent sample pair $(\theta_i,s_{\sigma(j)})\succ  (\theta_i, s_{\sigma(j+1)})$. Therefore, we define the preference loss based on the preference: 


\vspace{-5pt}
\begin{align}
\label{eq:lp}
\textstyle L_p(\phi) = -\frac{1}{N}\sum_{i=1}^{N} \sum_{j=1}^{M^P-1} &[ \mu_{i,j} \log(p_{i,j} ) + \\ \nonumber
&(1 - \mu_{i,j}) \log(1 - p_{i,j} ) ],
\end{align}
where the preference labels $\mu_{i,j}= 1$ if $(\theta_i,s_{\sigma(j)}) \succ  (\theta_i, s_{\sigma(j+1)})$, $0.5$ otherwise.

If the model outputs an incorrect order, such as the higher $\hat{r}_{i,\sigma(2)}$ leading $\hat{r}_{i,\sigma(1)}<\hat{r}_{i,\sigma(2)}$ and $\hat{r}_{i,\sigma(2)}>\hat{r}_{i,\sigma(3)}$, although \( (\hat{r}_{i,\sigma(1)},\hat{r}_{i,\sigma(2)}) \) would be penalized by \(L_p(\phi)\), the error in \( (\hat{r}_{i,\sigma(2)},\hat{r}_{i,\sigma(3)}) \) might even reduce the value of \(L_p(\phi)\). To address this issue, we introduce a reward difference-based loss to penalize the incorrect output of $\hat{r}_{i,\sigma(2)}$ and to reinforce the correct order within the sequence:

\vspace{-5pt}
\begin{align}
\label{eq:ld}
\textstyle L_d(\phi) = \frac{1}{N}\sum_{i=1}^{N} \sum_{j=1}^{M^P-1} (\hat{r}_{i,\sigma(j)} - \hat{r}_{i,\sigma(j+1)} - \delta_{i,j})^2,
\end{align}
where \(\delta_{i,j}=r(\theta_i, s_{\sigma(j)}) - r(\theta_i, s_{\sigma(j+1)})\) is the target reward differences between adjacent strategies in the sequence. The loss can effectively deepen the relative preferences and improves training stability. To enhance the coordination between the loss functions, finally, our total loss is:
\begin{align}
\label{eq:losstotal}
L_{total}(\phi) =  \lambda_1 L_p(\phi) + \lambda_2 L_d(\phi),
\end{align}
\noindent where $\lambda_1$ and $\lambda_2$ are hyperparameters for different scenarios. The framework of our method is shown in Algorithm \ref{alg:2}.

\vspace{-4pt}
\section{Online Strategy Inference}
To overcome the potential prediction errors, reliability can be increased by taking the $\text{Top-}k$ output strategies as candidates. Given the parameterized preference model $R_\phi$, and the instance $\theta_i$, let $s_k$ be the set of the $k$ strategies corresponding to the $k$ largest outputs
\begin{align}
S_k = \{s_j \mid \hat{r}_{i,j} \in \text{Top-}k(\{\hat{r}_{i,1}, \hat{r}_{i,2}, \dots, \hat{r}_{i,M^P}\})\},
\end{align} 
where $\hat{r}_{i,j} \in \hat{R}_{i}$ is the output of preference model $R_\phi$. We select $S_k$ as the strategy candidates for the instance $\theta_i$, and evaluate the strategies $s_j \in S_k$ by solving the reduction model $s_j(\theta_i)$. And the $s_j$ with the lowest infeasibility $p(\theta_i,s_j)$ is selected as the target strategy. 
\begin{align}
\hat{s} = argmin_{s_j \in S_k} p(\theta_i,s_j).
\end{align} 

To solve the reduction model $s_j(\theta_i)$, for special types of problems such as MIQP (Mixed-Integer Quadratic Programming) and MILP, the linear system can be simplified based on the KKT optimality conditions \cite{BV2014}, further speeding up the solution time. The workflow in online stage is detailed in Figure \ref{fig_online}

\vspace{-2pt}
\section{Experiments}
In this section, we compare our proposed method with the learning-based methods and the commercial solvers on real-world datasets to validate the performance and efficiency.

\begin{figure*}[t]
\centering
\captionsetup{skip=2pt} 
\includegraphics[width=1\textwidth]{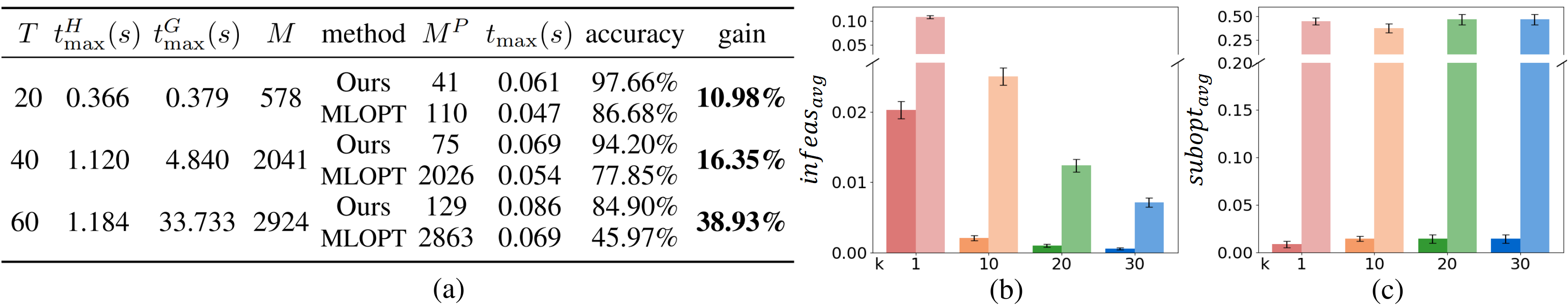}
\caption{Results on Fuel Cell Energy Management. (a) Performance under varying problem scale (larger scale as $T$ increasing), where $t_{\max}^H$ means the maximum computation time (in seconds) from Gurobi Heuristic and $t_{\max}^G$ is from Gurobi. (b) and (c) show the average infeasibility and suboptimality for our method (bar on the left) and MLOPT under $T$=60 and varying $k$.}
\label{fig_cell}
\end{figure*}

\begin{figure}[t]
\centering
\includegraphics[width=1\columnwidth]{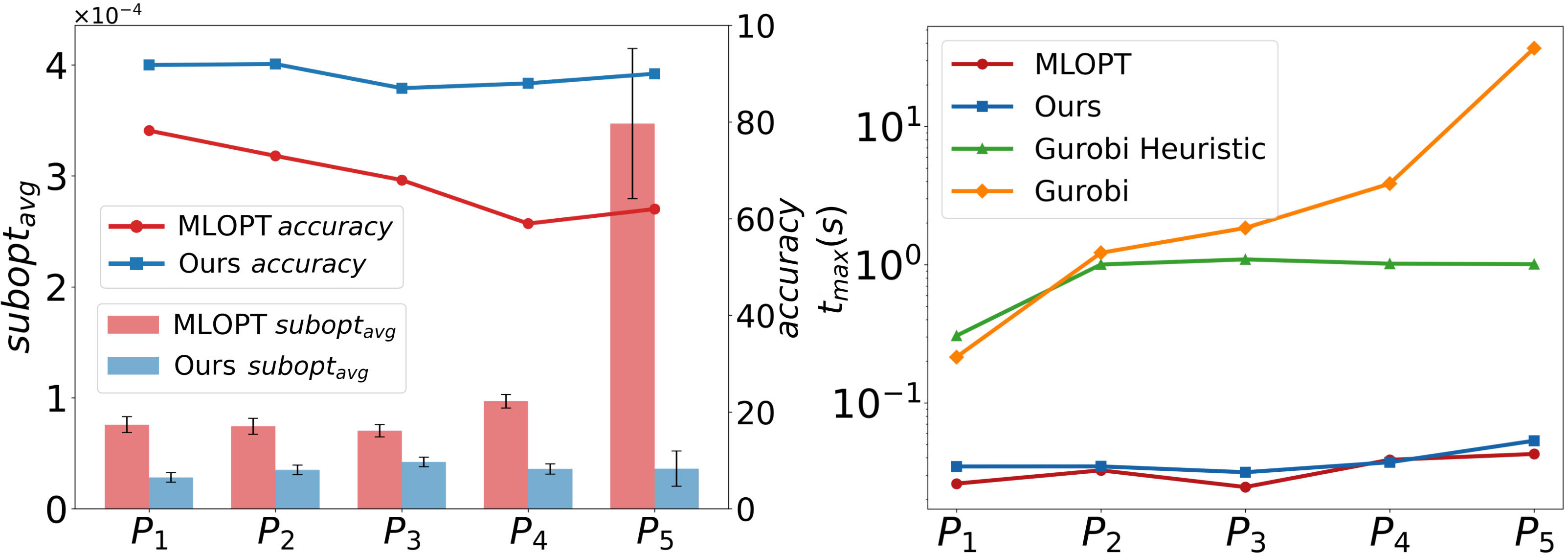}
\caption{Performance on Inventory Management Problems}
\label{resultlp}
\end{figure}

\textbf{Evaluation Metrics.}
We follow the same criteria for solving quality as in \cite{bertsimas2022online}, using the accuracy metric. We consider solutions to be accurate if their infeasibility and suboptimality are within a small tolerance. Given $N$ test  samples, the testing accuracy on this dataset is:
\vspace{-7pt}
\begin{align} \notag
accuracy = \frac{1}{N} \left| \{ \theta_{i} \mid p(\theta_i,\hat{s_j}) \leq \epsilon_1 \wedge d(\theta_i,\hat{s_j}) \leq \epsilon_2 \} \right|,   
\end{align}
where the tolerances $\epsilon_1$ for infeasibility and $\epsilon_2$ for suboptimality are both set to $1 \times 10^{-4}$.

\setlength{\itemsep}{1pt}     
\setlength{\topsep}{1pt}      
\setlength{\parsep}{1pt}      
\setlength{\parskip}{1pt}     
\textbf{Datasets.} We evaluate the performance through:
\begin{enumerate}
    \item \textit{MIPLIB} \cite{gleixner2021miplib}, six scenarios selected as in \cite{BertsimasK23}, the real-world MILP problems with varying scales and solving difficulties.
    \item \textit{Fuel Cell Energy Management Problem} \cite{frick2015embedded}, treated as the primary evaluation scenario by \cite{bertsimas2022online}. Its scale can be increased by increasing $T$ for deeper analysis.
    \item \textit{Inventory Management Problem}, five large-scale (average number of 100,000 constraints) real-world industrial problems from a company's real supply chain scenarios.
\end{enumerate}

\textbf{Baselines.}
We compare our method with:
\begin{enumerate}
    \item \textit{Gurobi} \cite{gurobi2021gurobi}, the advanced commercial solver. To make the comparison as fair as possible, we run Gurobi with ``warm-start'' enabled, that reuses the solution obtained from previous parameters.
    \item \textit{Gurobi Heuristic}, Gurobi's ``heuristic'' mode, a very fast heuristic algorithm with time limit of one second.
    \item \textit{MLOPT} \cite{BertsimasK23}, a model reduction based method that is the most applicable learning-based method in our scenario.
    \item \textit{Predict and Search} \cite{HanYCZZW0L23}, a solution-prediction based method with initial variable prediction and further neighborhood searching.
\end{enumerate}

In the datasets 2 and 3, the number of integer variables is small, and the Predict and Search method offers limited acceleration, making it less applicable. Thus, we do not compare the Predict and Search method in the datasets 2 and 3. For Datasets 1 and 3, we used 1000 samples for training and 500 for testing; for Dataset 2, 10,000 samples were used for training and 1000 for testing. The experiments are conducted on 32-core Intel CPUs (2.16 GHz), and the detailed setups and dataset formulations are presented in the Appendix.

\textbf{Evaluation on MIPLIB.}
Figure~\ref{radar} shows the performance on MIPLIB, from which we observe that: 1) Our method performs nearly as well as Gurobi on suboptimality and feasibility across most scenarios. 2) Our method shows a slight advantage in feasibility compared to MLOPT and a significant improvement in suboptimality. 3) Due to the complexity of the problem, heuristic methods struggle to provide high-quality solutions within the time limit, leading to poor performance across multiple metrics. 4) The Predict and Search method, which only predicts partial integer variables, performs well in terms of feasibility, but its search process can cause it to get trapped in local optimal, and the search sub-task is more time-consuming compared to our method.

\textbf{Evaluation on Fuel Cell Energy Management problem.}
As shown in Figure~\ref{fig_cell}, our method achieves an improvement of nearly 39\% in accuracy when the problem is the most complex. In 4(a), $M$ represents the number of strategies and $M^P$ represents the strategies after pruning. Our pruning method reduces the number of strategies much more than the pruning of MLOPT, especially as the problem size increases. The pruning of MLOPT performs poorly due to the difficulty in reassigning labels with some key strategies discarded. From 4(b) and 4(c), it can be observed that as $k$ increases, the performance of the model improves, and our method yields better outcomes than MLOPT in both key metrics under the same $k$ because: 1) The reduced strategy space increases the likelihood of finding effective strategies within the $\text{Top-}k$. 2) The preference model can evaluate the performance of all strategies for the given instance, allowing it to identify a more reasonable set of $k$-candidate strategies.

\textbf{Evaluation on Inventory Management problem.}
Figure~\ref{resultlp} shows the results on this scenario, where $P_1$ to $P_5$ denote five increasing model sizes. We observe that in this scenario, our method can maintain feasibility across all instances, thus only the suboptimality and accuracy metrics are reported. As the problem size increases, our method consistently achieves low average suboptimality, with accuracy remaining stable at around 90\%. On the largest scale problem $P_5$ which has more than 270,000 constraints, our method also outperforms MLOPT by approximately 30\%.

\textbf{Computation Time.}
Among the three datasets, our method, requiring only online inference and solving the linear system of reduced model, is significantly faster than Gurobi and heuristic algorithms. Compared with Gurobi, the computation time is improved by three orders of magnitude, and remains relatively stable as the problem size grows ($T$). Compared with MLOPT, because of the similar workflow, there is only a tiny difference in $t_{\max}$ coming from the size of the neural network under the same value of $k$. Compared with Predict and Search on the MIPLIB dataset (Figure~\ref{radar}), our reduction-prediction method is more time-efficient because the searching within high dimensionality neighbourhood is relatively time-consuming. Note that since the $\text{Top-}k$ computation can be executed in parallel, our method could even be an order of magnitude faster.

\vspace{-7pt}
\section{Conclusion}
\vspace{-5pt}
This paper proposes a preference-based model reduction (i.e., strategy) learning for fast and interpretable MILP solving. There are two challenges for strategy learning: 1) how to generate sufficient strategies that are useful for strategy learning, and 2) how to integrate the performance information of all candidate strategies to improve strategy learning. This paper first introduces the {\sc SetCover} technique to find a minimal set of strategies that can be applied to all MILP instances. Furthermore, the preference information of these available strategies on instances and the attention mechanism are integrated to improve the learning capacity. Simulations on real-world MILP problems show that the proposed method has a significant improvement in solving time and the accuracy of the output solutions.

\section{Acknowledgments}
This research is supported in part by Key Research and Development Projects in Jiangsu Province under Grant BE2021001-2, in part by the National Natural Science Foundation of China 62476121,62202100 and 61806053, and in part by the Jiangsu Provincial Double-Innovation Doctor Program No. JSSCBS20220077.

\bibliography{aaai25}

\clearpage
\section{Appendix}

\textbf{Dataset and Experimental Setup Supplement.}
We provide a detailed introduction of the datasets. Datasets 1 and 2 are publicly available, and our settings are largely consistent with previous works. Dataset 3 represents five problem models currently employed by a company, and we briefly describe its scenario due to commercial privacy concerns. However, we have anonymized the data and uploaded a de-identified sample of the model's LP relaxation as a demo in the supplementary material.

\textbf{Dataset 1} is from the MIPLIB library \cite{gleixner2021miplib}, and given in the $(A_{eq},b_{eq},A_{ineq},b_{ineq},c,lb,ub,I)$ format. The objective is to minimize $c^Tx$ over the feasible region $\{x : A_{ineq}x \leq b_{ineq},A_{eq}x = b_{eq},lb \leq x \leq ub\}$, where $I$ is the set of indices for the integer decision variables. We select the key parameters $\bar{\theta}$ for instance generation, treating the original data $\bar{\theta}$ as the center of the ball and generate parameters $\{\bar{\theta}_i\}_{i=1}^N$ uniformly from the ball $B(\bar{\theta},r)$. The ranges of $\bar{\theta}$ and $r$ in our experiments are broader than those in \cite{BertsimasK23} for more challenging applications. The varying parameters $\bar{\theta}$ mainly include $b_{eq}, b_{ineq},$ or $c$ and we ensure that none of their entries contain zeros to avoid unreasonable disturbances. 

\textbf{Dataset 2} represents the Fuel Cell Energy Management scenario. Switching fuel cells on and off reduces battery lifespan and increases energy loss; therefore, they are often paired with energy storage devices (such as supercapacitors) to reduce switching frequency during rapid transients. In this scenario, the objective is to control the energy balance between the storage device and the fuel cell to match the required power demand \cite{frick2015embedded}. The goal is to minimize energy loss while maintaining the switching frequency of the fuel cell within an acceptable range to prevent lifespan degradation. The model is as follows:

\begin{equation}
\label{eq_cell}
\begin{aligned}
&\min_{\mathbf{P}, \mathbf{z}} \sum_{t=0}^{T-1} f(P_t, z_t) = \sum_{t=0}^{T-1} (\alpha P_t^2 + \beta P_t + \gamma z_t) \\
&\text{s.t.} \\
&E_{t+1} = E_t + \tau (P_t - P_t^{\text{load}}), \quad t = 0, \ldots, T \\
&E^{\min} \leq E_t \leq E^{\max}, \quad t = 0, \ldots, T-1 \\
&0 \leq P_t \leq z_t P^{\max}, \quad t = 0, \ldots, T \\
&z_{t+1} = z_t + \omega_t, \quad t = 0, \ldots, T \\
&s_{t+1} = s_t + d_t - d_{t-T}, \quad t = 0, \ldots, T-1 \\
&s_t \leq n^{\text{sw}}, \quad t = 0, \ldots, T \\
&G(\omega_t, z_t, d_t) \leq h, \quad t = 0, \ldots, T \\
&E_0 = E_{\text{init}}, \quad z_0 = z_{\text{init}}, \quad s_0 = s_{\text{init}} \\
&z_t \in \{0, 1\}, \quad d_t \in \{0, 1\}, \quad \omega_t \in \{-1, 0, 1\}.
\end{aligned}
\end{equation}
In the objective function, $z_t \in {0,1}$ represents the ON-OFF state of the cell, and fuel consumption is given by $\alpha P_t^2 + \beta P_t + \gamma z_t$ ($\alpha,\beta,\gamma > 0$) when the battery is on ($z_t = 1$), where $P_t \in [0,P^{\max}]$ is the power provided by the fuel cell. This problem is a type of Mixed Integer Quadratic Program (MIQP). $E_t \in [E^{\min},E^{\max}]$ denotes the energy stored, and $\tau > 0$ is the sampling time. $d_t \in {0,1}$ determines whether the cell switches at time $t$, and $s_t$ is the number of switchings that have occurred up to time $t$. The complexity of the problem model is related to the time step $T$. For a fair comparison, we maintain consistency with the problem parameters ($\bar{\theta}$) and sampling methods ($B(\bar{\theta},r)$) used by \cite{bertsimas2022online} in this scenario. The problem parameters are $\theta=(E_{init},z_{init},s_{init},d^{past},P^{load})$, where $d^{past}=(d_{-T},...,d_{-1})$ and $P^{load} = (P_0^{load},...,P_{T-1}^{load})$. Further information can be found in \cite{frick2015embedded}.

\textbf{Dataset 3} is a type of Inventory Management Problem. The primary constraints in this scenario include the daily balance of goods entering and leaving the warehouse, limitations on the number of products and replacement parts, and demand fulfillment constraints. The objective function aims to minimize storage and transport costs while meeting demand. In practical applications, the key varying parameters $\bar{\theta}$ include the initial storage quantities and the costs in the objective function $c$ and constraints $A, b$. This dataset represents a large-scale, real-world supply chain application featuring hundreds of thousands of constraints and variables.

For experimental settings, the AdamW was used as the optimizer, with an initial learning rate ranging between 0.0001 and 0.001, depending on the problem size. This learning rate was linearly decayed by a factor of 0.9 every 10 epochs, continuing until 100 epochs were completed. The model was trained with a batch size of 128. The parameter $ \lambda_1$ in (\ref{eq:losstotal}), which coordinates $L_p$ and $L_d$ was set between 0.8 and 0.9, depending on the problem scenario. Further parameters and detailed settings can be found in the submitted code.

\begin{figure*}[t]
\centering
\includegraphics[width=1\textwidth]{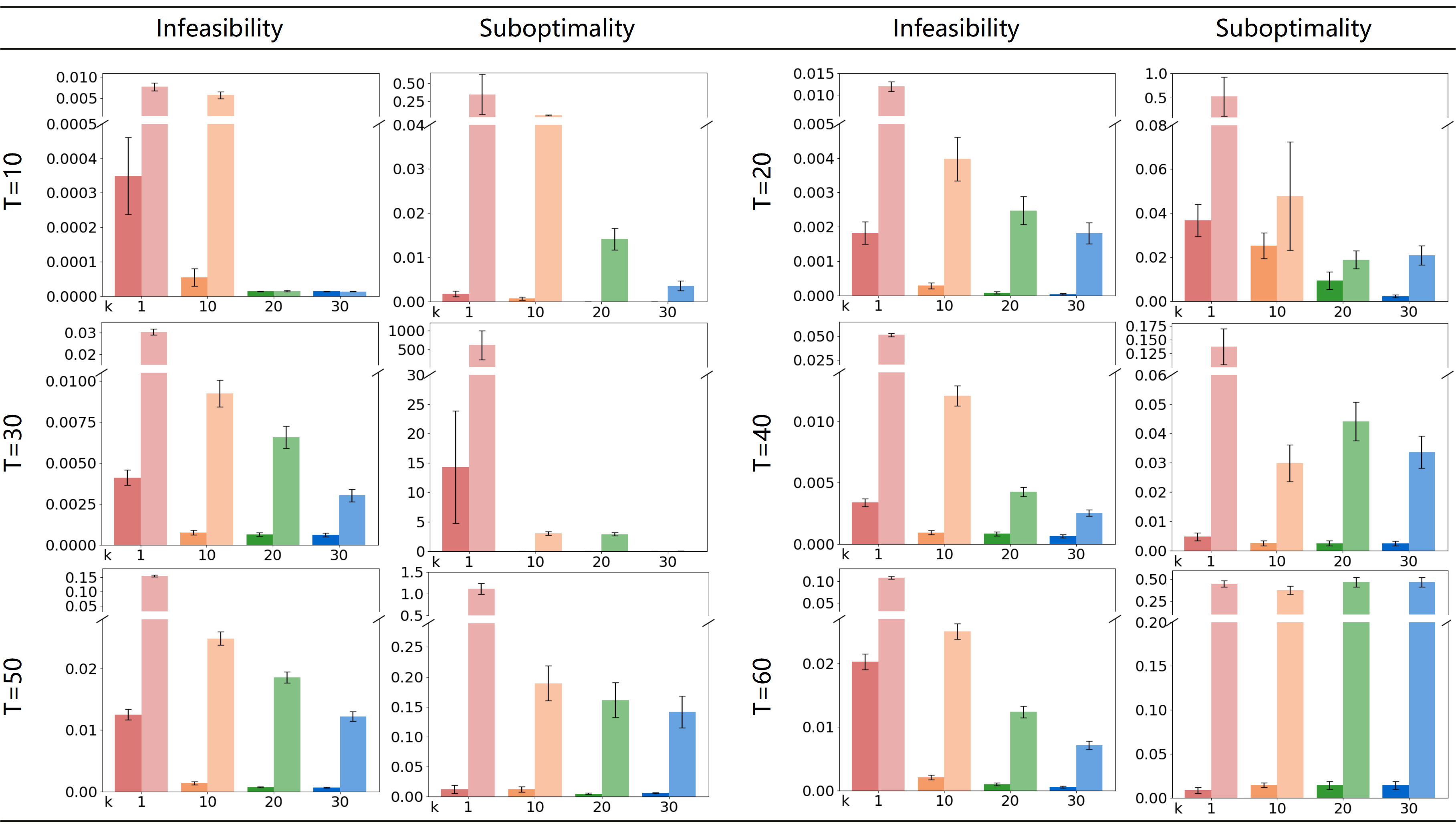}
\caption{The infeasibility and suboptimality metrics of our method (bar on the left) and MLOPT under varying $T$ and $k$}
\label{fig_cellall}
\end{figure*}

\textbf{Supplementary Results on the Fuel Cell Energy Management Problem.}
Due to space limitations in the main text, we only present the performance of our method on a subset of scenarios from Dataset 2. The complete results are shown in Figure \ref{fig_cellall}. As can be seen, compared to MLOPT, our method consistently demonstrates superiority in infeasibility and suboptimality metrics across all scenarios at the same $k$ values. Moreover, our method exhibits lower standard errors, indicating greater consistency across these metrics.

\begin{figure}[t]
\centering
\includegraphics[width=1\columnwidth]{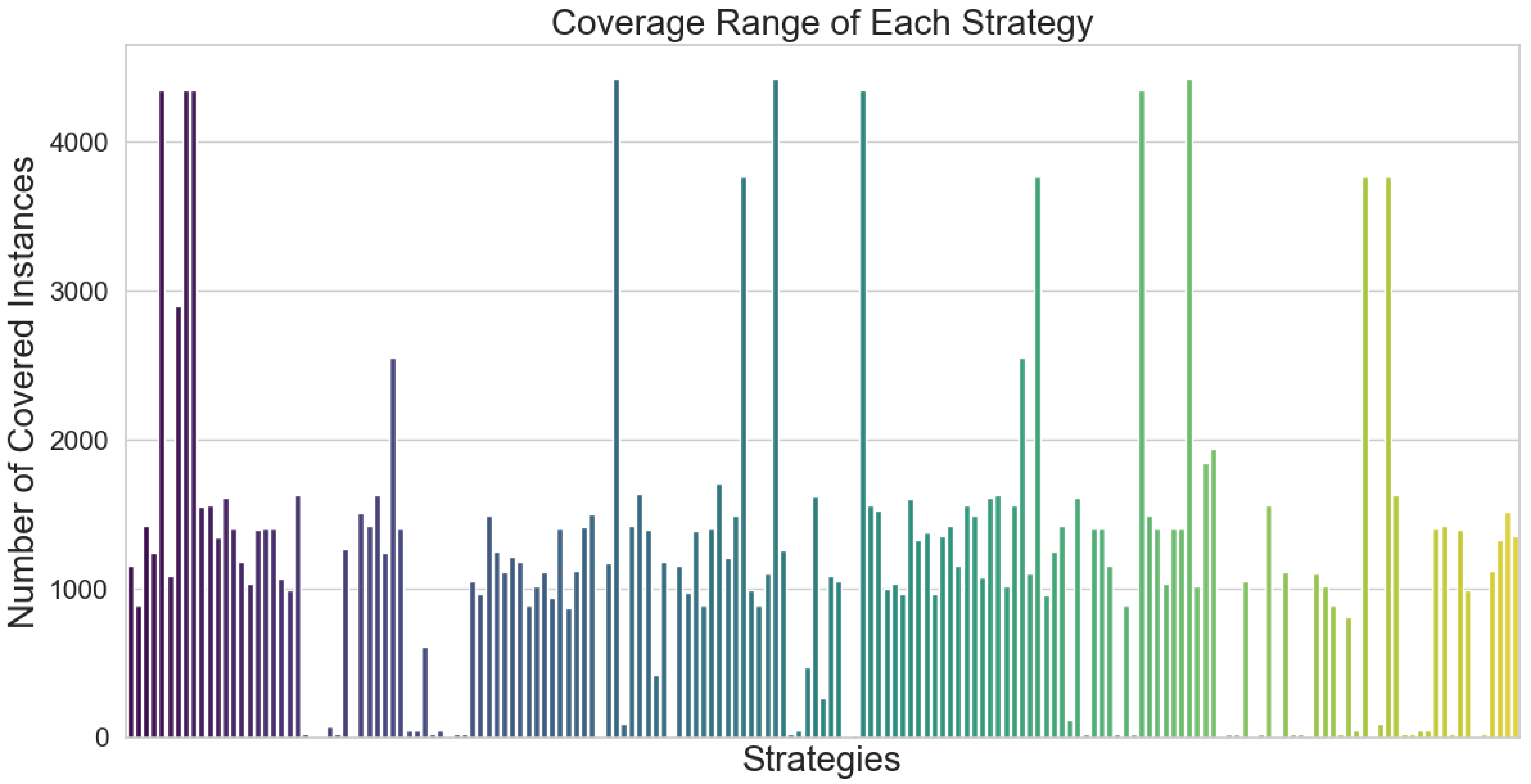}
\caption{The number of instances covered by the strategies}
\label{fig_cover_statistics}
\end{figure}

\textbf{Coverage Range of Strategies.}
\cite{bertsimas2022online} proposed a heuristic pruning method that removes low-frequency strategies based on their occurrence frequency in the training set $\left \{ \theta_{i}, s^{*}(\theta_{i})\right \}_{i \in [N]} $. This method only counts each instance's labeled strategy when calculating frequency, without considering other potential strategies applicable to the instances, which may lead to incorrect strategy reduction. Moreover, if there are no matching strategies among the unpruned strategies for instances whose labeled strategies have been pruned, the number of pruned strategies will decrease, resulting in incomplete and inefficient pruning.

Using the problem dataset 2 with the $T=10$ scenario as an example, we sample 10,000 instances to generate strategies and calculate the number of instances applicable to each strategy. As shown in Figure \ref{fig_cover_statistics}, many strategies cover more than thirty percent of the samples, with each strategy covering on average more than ten percent of the samples. It can be observed that strategies can be applied to multiple instances, and the number is much larger than their occurrence as labels in the dataset. This is why we explore a more thorough pruning method using {\sc SetCover}. The workflow of our algorithm is shown in Algorithm \ref{alg:setcover}.

\begin{algorithm}[tb]
\caption{Strategy Pruning on Bipartite Graph}
\label{alg:setcover}
\textbf{Input}: $V_\theta$: Set of instance nodes, $V_s$: Set of strategy nodes, $E$: Set of edges\\
\textbf{Output}: $V_s^{'}$: The set of strategy nodes covering all instance nodes
\begin{algorithmic}[1] 
\STATE $U \gets V_\theta$, Initialize uncovered instance nodes
\STATE $V_s^{'} \gets \emptyset$, Initialize selected strategy nodes
\WHILE{$U \neq \emptyset$}
\STATE $v_s^* \gets \arg\max_{v_s \in V_s} |\{v_{\theta} \in U \mid (v_{\theta}, v_s) \in E\}|$, Select the strategy node covering most uncovered instance nodes
\STATE $V_s^{'} \gets V_s^{'} \cup \{v_s^*\}$\\
\STATE $U \gets U \setminus \{v_\theta \in U \mid (v_\theta, v_s^*) \in E\}$\\
\ENDWHILE
\STATE \textbf{return} $V_s^{'}$
\end{algorithmic}
\end{algorithm}

\textbf{Experiments on Learning Objective.}
The reward model $R_\phi$ can be trained by directly fitting the rewards as object:

\begin{align}
L_r(\phi) = \frac{1}{M} \sum_{i=1}^{N} \sum_{j=1}^{M} (r(\theta_i, s_j) - \hat{r}_{i,j})^2,
\end{align}
where $\hat{r}_{i,j} \in \hat{R}_i$ is the predicted reward for strategy $s_j$ applied in instance $\theta_i$ by the model $R_\phi$. We do experiments to compare the performance between this Reward-Fitting learning objective (RF in short) and our preference-based learning objective as in Eq. (\ref{eq:losstotal}). We found that on the two larger datasets, Datasets 1 and 3, fitting rewards in higher dimensions became increasingly difficult, leading to significant fluctuations in the results. Therefore, we report the performance of the reward-fitting method on Dataset 2. We used the same pruning method, controlled for the same network structure, and conducted parameter tuning within the same range to compare the model's performance at $k=1$ across different problem sizes ($T$).

\setlength{\tabcolsep}{1pt}
\begin{table*}[t]
\centering
\caption{Performance of our method versus Reward Fitting (RF) on Fuel Cell Energy Management Problem when $k=1$}
\captionsetup{skip=2pt}
\label{tab:RF}
\begin{tabular}{ccccccccccccc}
\toprule \rule{0pt}{2ex}
       & \multicolumn{4}{c}{$T=10$}                                                & \multicolumn{4}{c}{$T=20$}                                                 & \multicolumn{4}{c}{$T=30$}                                                \\ \midrule \rule{0pt}{2ex} 
method & accuracy & $infeas_{avg}$ & $subopt_{avg}$ & gain                             & accuracy & $infeas_{avg}$ & $subopt_{avg}$ & gain                              & accuracy & $infeas_{avg}$ & $subopt_{avg}$ & gain                             \\ \midrule \rule{0pt}{2ex}
Ours   & 95.65\%  & 0.00035     & 0.00176     & \multirow{2}{*}{\textbf{4.58\%}} & 83.04\%  & 0.00182     & 0.03676     & \multirow{2}{*}{\textbf{10.60\%}} & 73.88\%  & 0.00409     & 14.33399    & \multirow{2}{*}{\textbf{9.38\%}} \\
RF     & 91.07\%  & 0.00680     & 0.00836     &                                  & 72.43\%  & 0.00685     & 0.03883     &                                   & 64.51\%  & 0.00705     & 0.04398     &                                  \\ \toprule \rule{0pt}{2ex}
       & \multicolumn{4}{c}{$T=40$}                                                & \multicolumn{4}{c}{$T=50$}                                                 & \multicolumn{4}{c}{$T=60$}                                                \\ \midrule \rule{0pt}{2ex} 
method & accuracy & $infeas_{avg}$ & $subopt_{avg}$ & gain                             & accuracy & $infeas_{avg}$ & $subopt_{avg}$ & gain                              & accuracy & $infeas_{avg}$ & $subopt_{avg}$ & gain                             \\ \midrule \rule{0pt}{2ex}
Ours   & 73.21\%  & 0.00338     & 0.00477     & \multirow{2}{*}{\textbf{9.60\%}} & 56.14\%  & 0.01250     & 0.01210     & \multirow{2}{*}{\textbf{14.29\%}} & 39.71\%  & 0.02029     & 0.00865     & \multirow{2}{*}{\textbf{6.82\%}} \\
RF     & 63.62\%  & 0.00740     & 0.01307     &                                  & 41.85\%  & 0.02385     & 0.01668     &                                   & 32.90\%  & 0.01387     & 0.04394     &                                  \\ \bottomrule
\end{tabular}
\end{table*}

\setlength{\tabcolsep}{1pt}
\begin{table*}[t]
\centering
\caption{Evaluation on the ranking sampling on Fuel Cell Energy Management Problem when $k=1$}
\captionsetup{skip=2pt}
\label{tab:CN2}
\begin{tabular}{ccccccccccccc}
\toprule \rule{0pt}{2ex}
       & \multicolumn{4}{c}{$T=10$}                                                & \multicolumn{4}{c}{$T=20$}                                                 & \multicolumn{4}{c}{$T=30$}                                                \\ \midrule \rule{0pt}{2ex} 
method & accuracy & $infeas_{avg}$ & $subopt_{avg}$ & gain                             & accuracy & $infeas_{avg}$ & $subopt_{avg}$ & gain                              & accuracy & $infeas_{avg}$ & $subopt_{avg}$ & gain                             \\ \midrule \rule{0pt}{2ex}
Ours   & 95.65\%  & 0.00035     & 0.00176     & \multirow{2}{*}{\textbf{5.08\%}} & 83.04\%  & 0.00182     & 0.03676     & \multirow{2}{*}{\textbf{8.48\%}} & 73.88\%  & 0.00409     & 14.33399    & \multirow{2}{*}{\textbf{6.70\%}} \\
NR     & 89.84\%  & 0.00119      & 0.00069     &                                  & 74.55\%  & 0.00315     & 0.01899     &                                   & 67.19\%  & 0.00639     & 0.02675     &                                  \\ \toprule \rule{0pt}{2ex}
       & \multicolumn{4}{c}{$T=40$}                                                & \multicolumn{4}{c}{$T=50$}                                                 & \multicolumn{4}{c}{$T=60$}                                                \\ \midrule \rule{0pt}{2ex} 
method & accuracy & $infeas_{avg}$ & $subopt_{avg}$ & gain                             & accuracy & $infeas_{avg}$ & $subopt_{avg}$ & gain                              & accuracy & $infeas_{avg}$ & $subopt_{avg}$ & gain                             \\ \midrule \rule{0pt}{2ex}
Ours   & 73.21\%  & 0.00338     & 0.00477     & \multirow{2}{*}{\textbf{9.49\%}} & 56.14\%  & 0.01250     & 0.01210     & \multirow{2}{*}{\textbf{5.25\%}} & 39.71\%  & 0.02029     & 0.00865     & \multirow{2}{*}{\textbf{13.80\%}} \\
NR     & 63.73\%  & 0.00623     & 0.01223     &                                  & 50.89\%  & 0.01184     & 0.00366     &                                   & 25.91\%  & 0.01411     & 0.00581     &                                  \\ \bottomrule
\end{tabular}
\end{table*}

As shown in Table \ref{tab:RF}, our method achieved an average accuracy improvement of approximately 9\% across all scenarios compared to RF. Additionally, our method consistently outperformed RF in terms of average infeasibility and suboptimality in almost all scenarios. Although the average suboptimality of our method was higher at $T=30$, its feasibility and accuracy remained superior to RF. This was due to a few isolated instances with unusually high suboptimality. In fact, Reward Fitting still outperformed MLOPT, indicating that introducing rewards to evaluate the strategies is effective and allows the model to leverage more strategy-related information.

The reason RF performs worse than our preference-based method is likely because, compared to the preference objective, fitting specific scores in higher dimensions is prone to prediction errors. Since the data has been normalized, even minor errors can lead to misjudgments in identifying the optimal strategy. Furthermore, the reward differences among the Top-$k$ strategies might be minimal, which further complicates identifying the best strategy. In contrast, our training approach emphasizes relative preferences among strategies, aiming to maintain an ordered sequence and the differences between strategies within the sequence. This approach strengthens the position of the best strategy at the top of the sequence, amplifying the differences between strategies, and thus has a clear advantage in identifying the most preferred strategy for a given instance.

\textbf{Samples for Preference Learning.}
Traditional preference learning trains models using the approach described in Eq. (\ref{eq:lp}) with $\binom{M^{P}}{2}$ sample pairs. Our method improves this sampling approach by incorporating strategy ranking and reward difference fitting. We quantitatively compared these two sampling methods through experiments. Sampling all possible preference pairs is prohibitively expensive and impractical in real-world scenarios. To ensure fairness, we trained the models using an approximately equal number of samples, randomly selecting more than $n$ pairs from all available preference pairs while ensuring that every strategy appears in at least one preference pair. We conducted these experiments on Dataset 2.

Table \ref{tab:CN2} shows the results of different methods, where "NR" (no ranking) represents the original sampling method without ranking. As shown in the table, our method achieved an average accuracy improvement of around 8\% across various problem sizes. Moreover, our method generally outperformed in terms of average infeasibility and suboptimality in most scenarios. In a few cases where the average metrics were slightly inferior, the overall accuracy still favored our approach, indicating that the differences were due to the higher variance in a small number of instances. Overall, our method demonstrates that by utilizing sequence preference sampling to train the model and reinforcing preference differences through the reward difference loss, it better leverages the preference relationships between strategies. Additionally, our method uses the minimum number of samples necessary for preference training, significantly reducing training costs, which makes it feasible to apply preference learning to larger-scale problems.

\end{document}